%% file: main.tex
\documentclass[10pt,twocolumn,letterpaper]{article}

 \usepackage{iccv}             
\usepackage{times}
\usepackage{epsfig}
\usepackage{graphicx}
\usepackage{amsmath}
\usepackage{amssymb}
\usepackage{animate}
\usepackage{booktabs}
\usepackage{multirow}
\usepackage{multicol}
\usepackage{bbm}
\usepackage{xcolor}
\usepackage{colortbl, booktabs}
\usepackage{makecell}
\newcommand\blfootnote[1]{%
  \begingroup
  \renewcommand\thefootnote{}\footnote{#1}%
  \addtocounter{footnote}{-1}%
  \endgroup
}

\definecolor{iccvblue}{rgb}{0.21,0.49,0.74}
\usepackage[pagebackref,breaklinks,colorlinks,allcolors=iccvblue]{hyperref}

\def\paperID{***} % *** Enter the Paper ID here
\def\confName{****}
\def\confYear{****}

\title{From Trial to Triumph: Advancing Long Video Understanding via Visual Context Sample Scaling and Self-reward Alignment}

\author{Yucheng Suo$^{1*}$, Fan Ma$^{1}$, Linchao Zhu$^{1}$, Tianyi Wang$^{2}$, Fengyun Rao$^{2}$, Yi Yang$^{1\dag}$\\
$^{1}$ReLER, CCAI, Zhejiang University ~~ $^{2}$Wechat Vision, Tencent Inc. \\
}

\begin{document}
\maketitle

\input{sec/0_abstract}    
\blfootnote{$^{*}$Work done during internship at Wechat Vision}  \blfootnote{$^{\dag}$ Corresponding author}
\input{sec/1_intro}
\input{sec/2_related}

\input{sec/3_preliminary}
\input{sec/4_method}
\input{sec/5_experiments}
\input{sec/6_conclusion}
{
    \small
    \bibliographystyle{ieeenat_fullname}
    \bibliography{main}
}

\end{document}

%% file: sec/0_abstract.tex
\begin{abstract}
Multi-modal Large language models (MLLMs) show remarkable ability in video understanding. Nevertheless, understanding long videos remains challenging as the models can only process a finite number of frames in a single inference, potentially omitting crucial visual information. To address the challenge, we propose generating multiple predictions through visual context sampling, followed by a scoring mechanism to select the final prediction. Specifically, we devise a bin-wise sampling strategy that enables MLLMs to generate diverse answers based on various combinations of keyframes, thereby enriching the visual context. 
To determine the final prediction from the sampled answers, we employ a self-reward by linearly combining three scores: (1) a frequency score indicating the prevalence of each option, (2) a marginal confidence score reflecting the inter-intra sample certainty of MLLM predictions, and (3) a reasoning score for different question types, including clue-guided answering for global questions and temporal self-refocusing for local questions. The frequency score ensures robustness through majority correctness, the confidence-aligned score reflects prediction certainty, and the typed-reasoning score addresses cases with sparse key visual information using tailored strategies. Experiments show that this approach covers the correct answer for a high percentage of long video questions, on seven datasets show that our method improves the performance of three MLLMs.
\end{abstract}

%% file: sec/1_intro.tex
\section{Introduction}
\label{sec:intro}
As a medium for humans to perceive the real dynamic world, videos play a vital role in daily life applications \cite{miech19howto100m,caba2015activitynet,yang2024vidchapters} such as social communication, education, entertainment, etc.  Long video understanding poses significant challenges in the multi-modal research field, as it necessitates both spatial and temporal comprehension and reasoning based on a limited number of frames from potentially hour-long videos \cite{wang2024lvbench}. The recent advancements in Multi-modal Large Language Models (MLLMs) \cite{achiam2023gpt,team2023gemini,team2024gemini} improve the zero-shot video understanding accuracy by integrating spatial-temporal visual embeddings with the Large Language Models (LLMs) \cite{dubey2024llama,bai2023qwen} and fine-tuning on video-text instruction-tuning datasets. However, MLLMs encounter several common challenges. First, the training procedure only samples partial frames from the entire video, potentially missing fine-grained details. Second, the disparity in length between training and testing videos presents a length extrapolation challenge, while directly increasing the number of inference frames escalates memory usage.
\begin{figure}[t]
\begin{center}
\includegraphics[width=0.98\linewidth]{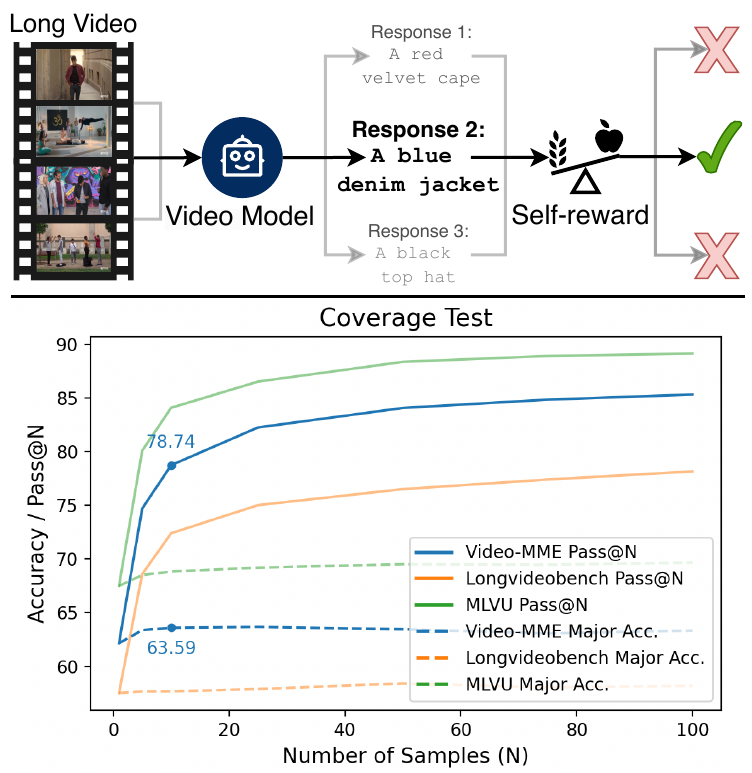}
\end{center}
\vspace{-.25in}
   \caption{\textbf{Pipeline and our motivation.} Our pipeline generates multiple predictions for long video questions and selects the best prediction using a self-reward score. This approach is inspired by a coverage test, which demonstrates that Pass@10 exceeds the accuracy of majority answer by $15\%$ using the Llava-Video model.}
\label{fig:first}
\vspace{-0.2in}
\end{figure}
Researchers investigate on efficiently handling longer sequences in MLLMs to deals with the challenges, which enables processing more frames simultaneously \cite{zhang2024long,xu2024slowfast,liu2024nvila,wang2024retake}. For instance, Video-XL \cite{shu2024video} clusters visual tokens based on semantic consistency and subsequently compresses the tokens within each cluster. Similarly, LongVU \cite{shen2024longvu} employs a text-aware pruning mechanism to reduce visual tokens efficiently. 
Another line of work focuses on test-time token reduction and window extension strategies. Slow-fast Llava \cite{xu2024slowfast} designs an inter-intra frame embedding pooling strategy. IG-VLM \cite{kim2024image} arranges sampled frames into grids of a single image, allowing image-based MLLMs to comprehend videos.
Despite these advancements, recent studies  \cite{wu2025longvideobench,zhang2024long} indicate that merely increasing the number of processed frames does not necessarily enhance performance on long videos. This is primarily because identifying key information within a vast array of visual tokens remains a significant challenge \cite{zhao2024needle}.

To address the aforementioned challenges, we draw inspiration from repeated sampling self-improvement strategies in LLM inference \cite{brown2024large,qiu2024treebon,chen2024language,dwaracherla2024efficient}.  These strategies often involve sampling various reasoning procedures for complex tasks.  
Our intuition is that long videos provide a dynamic visual context. However, each prediction is based on a finite number of frames, indicating that these frames may not fully capture the entire content of the video. By increasing the number of samples using different frames, the video model achieves a more comprehensive understanding of the long video. Therefore, we investigate how scaling up visual context sampling affects video large language model predictions.  
\cref{fig:first} illustrates an experiment where Llava-Video \cite{zhang2024video}, a state-of-the-art video large language model, performs multiple inferences based on 32 randomly sampled frames, assessing whether the correct answer appears among the predictions. Interestingly, the correct answer appears at least once in over $78\%$ questions in Video-MME among 10 predictions. When scaling up to 100 predictions, the model covers the correct answer in over $80\%$ questions. 

This empirical finding suggests a promising approach for enhancing long video understanding by selecting the optimal prediction from multiple inferences. To achieve this, we design a self-reward score comprising three components. The first component is a frequency score measuring the prevalence of each option, which leverages the inherent capabilities of the model and lays a foundation for precise prediction selection. The second component is a marginal confidence score, reflecting the certainty of each prediction by calculating the inter-intra sample logit difference. Considering different video questions require varying perception fields, we introduce an adaptive contextual voting score that integrates additional answers through a voting mechanism. For instance, the question ``What is the man in red doing?'' can be answered by examining a specific moment in the video, whereas the question ``What can be inferred from the video?'' requires a holistic understanding. For global questions, we employ video narration combined with clue-based question answering, leveraging the strong reasoning capabilities of an external text-only LLM to provide comprehensive insights. For local questions, we introduce a temporal self-refocus approach, leveraging the video model confidence toward various segments to locate the key event and generate additional answers.  
The proposed method does not require additional training, and each inference is performed on a fixed number of frames, thereby reducing computing resource demands.
Experiments conducted on seven long video benchmarks demonstrate that the proposed method consistently improves the performance of three MLLMs, with gains of up to $+4.28\%$ on Videomme and $+5.89\%$ on MLVU.

In essence, this paper makes the following contributions:
\begin{itemize}
    \item We discover that scaling visual frame sampling for video large language models achieves high coverage across different video questions. This leads to a new paradigm for improving long video understanding by generating various predictions and selecting the best option.
    \item We propose a hybrid score to select the optimal prediction by evaluating the frequency, marginal confidence, and integrating additional answers through voting.
    \item Experiments on seven long video benchmarks and three MLLMs validate the effectiveness of our method.
\end{itemize}

%% file: sec/2_related.tex
\section{Related Work}
\label{sec:related}

\begin{figure*}[t]
\begin{center}
\includegraphics[width=0.99\linewidth]{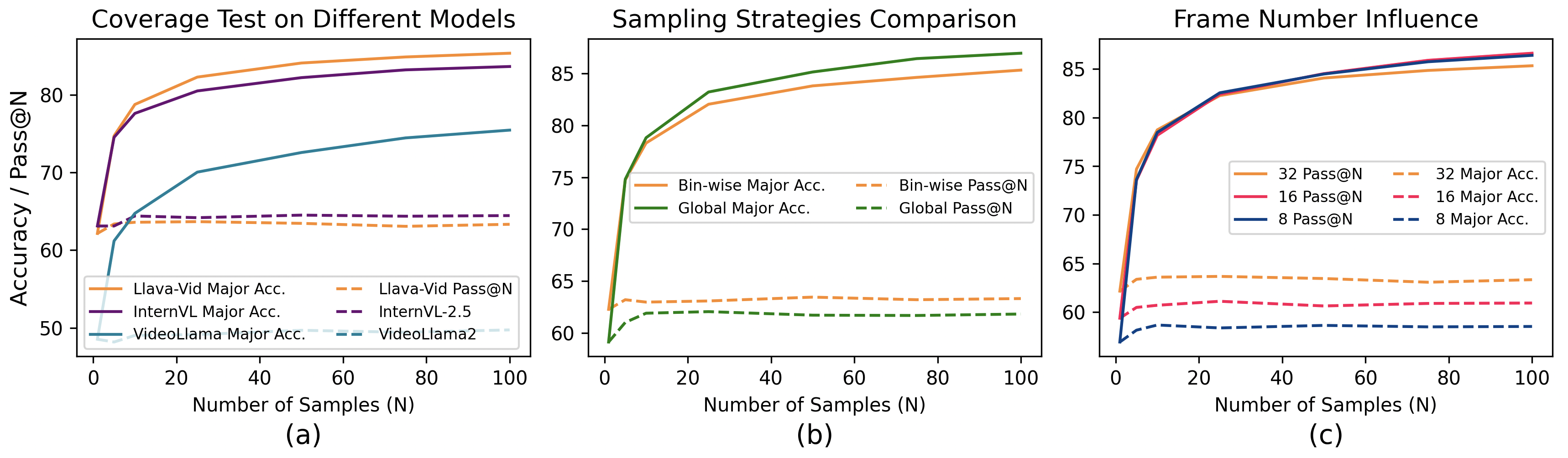}
\end{center}
\vspace{-.3in}
   \caption{\textbf{Scaling sampling experiments.} (a) Coverage test using different models, (b) comparison between global and bin-wise visual context sampling, (c) coverage test using fewer frames.}
\label{fig:coverage}
\vspace{-.1in}
\end{figure*}
\subsection{Multi-modal Large Language Models}
Recently, researchers have leveraged the reasoning ability of Large Language Models (LLMs) in the field of multi-modal understanding. BLIP \cite{li2022blip,li2023blip} is a representative work that uses q-former as a visual-to-text projector. LLava \cite{liu2023visual} uses a linear projector and fintunes on visual instruction tuning datasets \cite{zhang2024video,li2023mimic,li2023videochat,li2024mvbench} to boost the performance. Similar ideas are applied in the video tasks \cite{huang2024vtimellm,ma2024vista,li2024improving,NEURIPS2024_0ae94013,jin2024chat,li2024llama,maaz2023video,zhang2023video,luo2023valley,zohar2024apollo,liu2024oryx}. Common video large language models encode the video frames use CLIP \cite{radford2021learning} or SigLIP \cite{zhai2023sigmoid}, then apply a projector like MLP \cite{lin2023video, ataallah2024minigpt4,maaz2023video,zhang2024video}, Q-former \cite{zhang2023video}, etc. to match the large language model embedding space. Finally, the open source LLMs like Llama \cite{touvron2023llama}, Qwen \cite{bai2023qwen}, and InternLM \cite{cai2024internlm2}, etc. offer reasoning and text comprehension ability. 

Video tokens are numerous and exhibit redundancy, to effectively comprehend long videos, ST-LLM \cite{liu2024st} adopts dynamic masking in the projector to model global-local information, Chat-Univi \cite{jin2024chat} introduces DPC-KNN to cluster visual tokens, LongVLM \cite{weng2024longvlm} combines CLS token in frame-wise token pooling. NVILA \cite{liu2024nvila} uses scale-then-compress ideas to prune visual tokens. Video-XL \cite{shu2024video} introduces Visual Summarization Token to generate compact visual representations. VideoChat \cite{li2023videochat} and VideoLLaMA \cite{zhang2023video} on the other hand utilizes Q-Former to compress video representations. Another line of works \cite{zhang2024long,xue2024longvila,shen2025long} extrapolates language context window for long video understanding. In this work, we opt for three state-of-the-art Video LLMs, \ie Llava-Video \cite{zhang2024video}, InternVL2.5 \cite{chen2024expanding}, Videollama2 \cite{cheng2024videollama} 
, as the base models.
\subsection{Long Video Understanding}
Previous multi-modal benchmarks consider images \cite{singh2019towards,li2023seed,yu2023mm,liu2024mmbench,li2023evaluating} or short videos \cite{li2024mvbench,xiao2021next,yu2019activitynet,jang2017tgif,xu2017video}, testing the narrative\cite{yin2023survey,agrawal2019nocaps}, reasoning \cite{yue2024mmmu} and perception \cite{masry2022chartqa} ability on various aspects. Different from these tasks, long video understanding (LVU) is a challenging task since video large language models are only provided with visual context information \cite{wang2024videoagent}. To evaluate the zero-shot performance of video models, researchers curated benchmarks for LVU. Egoschema \cite{mangalam2024egoschema}, Cinepile \cite{rawal2024cinepile} and MovieChat-1K \cite{song2024moviechat} are early works, collecting long videos from egocentric videos and TV series respectively. Other representative works include Video-MME \cite{fu2024video}, LongVideoBench \cite{wu2025longvideobench}, and LVBench \cite{wang2024lvbench} containing longer videos and testing models on reasoning and perception ability. CG-Bench \cite{chen2024cg} offers visual clues in videos. Video-MMMU \cite{hu2025video} evaluates the knowledge-acquiring ability over professional videos. In this paper, we conduct experiments on seven long video understanding datasets, the shortest benchmark has an average video length of three minutes.

\subsection{Inference-time Self-improvements}
LLMs can benefit from increasing inference time computing through Chain-of-thought prompting \cite{wei2022chain,yao2024tree,wang2022self} and model-aided improvement \cite{liu2021dexperts,gao2023scaling,yuan2024self}. Progress have been made in application scenarios like coding \cite{roziere2023code,guo2024deepseek,li2023starcoder}, 	
arithmetic reasoning \cite{cobbe2021training,liu2024deepseek,achiam2023gpt,glm2024chatglm}. Repeated sampling \cite{hao2023reasoning} is another commonly used technique, Brown \etal \cite{brown2024large} investigates the scaling laws in language benchmarks and observes accuracy improves as sample number increases. Reward models \cite{cobbe2021training,lightman2023let,hosseini2024v,wang2024math} are deployed to score each sample and select Best-of-N \cite{irvine2023rewarding}. In the multi-modal field \cite{cheng2024vision}, works like Llava-CoT \cite{xu2024llava} and Mulberry \cite{yao2024mulberry} apply similar techniques to solve complex reasoning tasks. Inspired by recent works, this paper studies self-improvement in the video domain by scaling visual context input samples.

%% file: sec/3_preliminary.tex
\section{Scaling Visual Context Sampling}
\begin{figure*}[t]
\begin{center}
\includegraphics[width=0.995\linewidth]{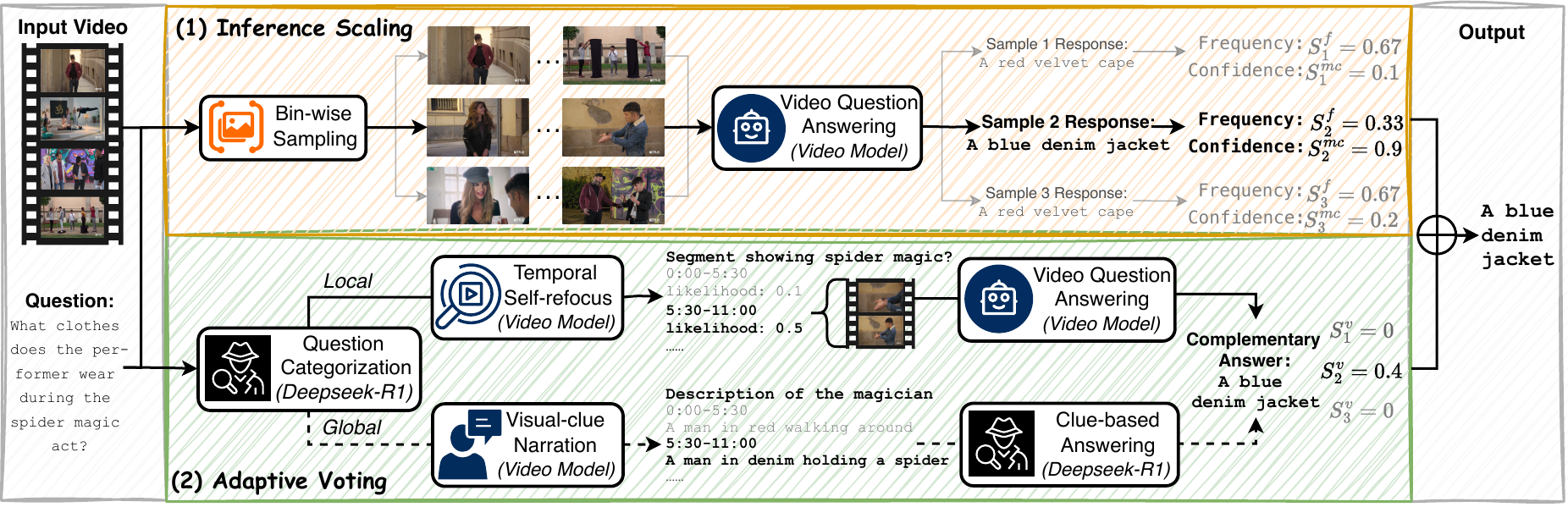}
\end{center}
\vspace{-.25in}
   \caption{\textbf{Method Overview.} Our method first adopts bin-wise visual context sampling to generate multiple predictions and then selects the final prediction with the largest self-reward score. The score consists of three components including the frequency score $\mathcal{S}^f$, marginal confidence score $\mathcal{S}^{mc}$, and a voting score $\mathcal{S}^v$ is calculated based on the complementary answer. }
\label{fig:coverage}
\vspace{-.15in}
\end{figure*}
During inference, multi-modal large language models usually uniformly sample a fixed number of frames as input for a long video, which potentially misses the key visual clues for the questions. To mitigate this issue, our idea is to random sample frames and make multiple predictions, thereby enlarging the perception field for the long videos. To investigate whether random sampling visual context influences the model prediction, we start with a coverage test shown in \cref{fig:first}. The definition of coverage, also known as Pass@N, is whether the correct answer $A$ appears at least once in the N sampled prediction set $\mathcal{P}$, which can be formulated as:
\begin{equation}
    \text{Pass@N} = \frac{\sum_{i=1}^{Q} \mathbbm{1}\{A_i \in \mathcal{P}_i\}}{Q}.
\end{equation}
$Q$ denotes the number of questions.
Coverage serves as the upper bound of the model given a fixed number of frames. \\
\textbf{Coverage test and a new paradigm.} \cref{fig:first} shows the Pass@N on Video-MME \cite{fu2024video}, Longvideobench \cite{wu2025longvideobench} and MLVU \cite{zhou2024mlvu} datasets using the Llava-Video \cite{zhang2024video} model. Additionally, \cref{fig:coverage}(a) presents the Pass@10 of three models, \ie Llava-Video, InternVL-2.5 \cite{chen2024expanding} and VideoLlama2 \cite{cheng2024videollama} on the Video-MME benchmark. Each prediction is based on 32 frames. Results show that different models consistently exhibit high coverage on long video benchmarks, suggesting that visual context sampling enables models to generate correct answers for most queries. 
Our findings offer a new paradigm to improve long video understanding by first sample predictions and then selecting the best one.
To this end, we test a straightforward method by selecting the most frequent answer, namely majority voting. While this method performed well, a $15\%$ gap remains compared to the Pass@10. Moreover, the gap is not reduced when increasing sample times. This suggests the need for an accurate selection strategy to enhance long video understanding.\\
\textbf{Details of the scaling sampling strategy.} We first investigate the impact of different sampling strategies on model performance. \cref{fig:coverage} (b) compares two sampling strategies, \ie, fully random sampling and bin-wise random sampling. Fully random sampling involves directly sampling 32 frames from a long video. In contrast, bin-wise random sampling uniformly divides the long video into 32 segments and randomly selects one frame from each segment. 
The merit of bin-wise sampling is that it ensures an even distribution of sample density across the timestamps, preventing the model from focusing solely on a specific part of the entire video. Results show that both the sampling strategy reaches $75\%$ Pass@5 and the Pass@100 surpasses $85\%$. Yet bin-wise sampling improves the majority accuracy by $+2\%$. Hence, we adopt bin-wise sampling as the default strategy. 

Additionally, \cref{fig:coverage}(c) illustrates how the number of frames in each sample affects Pass@N and majority accuracy. This experiment is motivated by the fact that fewer sample frames shorten sequence length and improve inference speed in video large language models. On the Video-MME benchmark, we evaluate the outcomes of randomly sampling 8 frames and 16 frames per prediction using the Llava-Video model. 
While the Pass@N are highly similar, the majority accuracy significantly decreases over $5\%$ when switching from 32 frames to 8 frames, implying increased model uncertainty. For better performance, the default sample frame number is 32 in later experiments.
Note that to evaluate the influence of visual context and follow the common setting in video models and benchmarks \cite{fu2024video,wu2025longvideobench,zhang2024lmmsevalrealitycheckevaluation,zhang2024video}, we set the LLM temperature and top-p to 0 for determined decoding. Moreover, for efficiency, the number of samples $N$ in the prediction set $\mathcal{P}$ is 10.

%% file: sec/4_method.tex
\section{Self-reward for Prediction Selection} 
Scaling up visual context sampling grants a feasible upper bound for MLLMs, yet selecting the correct prediction $\hat{p}$ from various answers remains challenging, as evidenced by the gap between majority accuracy and Pass@N. To bridge the gap, we propose a self-reward score for prediction selection. Our first intuition is that the frequency of verbalized prediction does not reflect model confidence, as all predictions are equally treated. Therefore, we propose a marginal-confidence score.
Furthermore, since different types of video questions require varying levels of video comprehension, an adaptive contextual voting score is introduced.

\subsection{Frequency and Marginal-confidence Scores}
The frequency score $\mathbf{S}^f_p$ measures the prevalence of the options among the prediction set $\mathcal{P}$ containing N samples. For a prediction $p \in \mathcal{P}$, $\mathbf{S}^f_p$ can be formulated as:
\begin{equation}
    \mathbf{S}^f_p = \frac{\sum_{j=1}^{N} \mathbbm{1}(\mathcal{P}_j = p)}{N}.
\end{equation}
Notably, only using $S^f$ for prediction selection is equivalent to the majority answer. The frequency score assigns equal weight to each prediction, overlooking the uncertainty introduced by variations in visual context quality. 

To estimate the confidence of each prediction, we utilize the logits of the first token predicted by the video large language model, since the first token probability is indicative of the model preference for a particular answer \cite{tu2024ranked}. 
However, directly employing the highest logit values ignores the intra-sample relative confidence. In other words, the absolute logit value does not indicate the confidence margin between the possible choices within each prediction. Additionally, since visual context quality varies across predictions, it is important to filter out low-confidence predictions that could affect the frequency-based selection process. Therefore, we introduce an inter-intra sample marginal confidence score $S^{mc}$ which can be formulated as:
\begin{equation}
    \mathbf{S}^{mc}_p = \max (l^{max}_j - l^{second}_j), \quad j \in \{k \mid \mathcal{P}_k = p\}.
\end{equation}
Here $l^{max}_j$ represents the logit value of the most-like token in prediction $\mathcal{P}_j$ and $l^{second}_j$ is the logit value of the token with the second high probability. The margin of the two logits depicts the internal relative confidence. $\mathbf{S}^{mc}$ picks the max value across samples with the same results, eliminating the influence of low-confidence predictions.
\subsection{Adaptive Contextual Voting Score}
\label{sec:type}
While combining $\mathbf{S}^f$ and $\mathbf{S}^{mc}$ yields a statistically optimal solution, it does not account for the varying degrees of visual context required by different long-video questions. Some questions necessitate a global understanding of the entire video, whereas others focus on specific local moments. 
To this end, we propose an adaptive contextual voting score $\mathbf{S}^v$ based on a complementary answer $A$. The following describes the procedure for generating $A$ and $\mathbf{S}^v$.\\
\textbf{Question categorization.} We first leverage the strong reasoning ability of an external Large Language model $\mathbf{M}$ to categorize the question into global or local questions, then use tailored strategies to generate the complementary answer $A$. Global reasoning questions typically involve understanding the overall theme, counting, ordering, or reverse questioning. While local questions like action recognition, and attribute perception require the video model to capture a specific moment to answer the question.\\
\textbf{Visual-clue narration for global reasoning questions.} 
Answering global questions requires the model to comprehend the entire video or reason across multiple moments in a long video. 
However, the reasoning process within each prediction is implicitly handled by the video model $\mathbf{V}$ based on a finite number of frames, which may include irrelevant visual context while lacking crucial global information.
To address this challenge, we propose using a surrogate large language model to explicitly accomplish the reasoning process using text generated by the video large language model. Specifically, the external language model $\mathbf{M}$ first analyses the question and identifies the key information necessary for answering it. The long video is then uniformly divided into $T$ segments, and the video large language model iteratively narrates all relevant visual clues $\mathcal{C}$ within each segment. In our implementation, $T$ is set to 8. Finally, the language model $\mathbf{M}$ produces a complementary answer $A$ by summarizing and reasoning over the extracted clues across segments. The procedure can be formulated as:
\begin{equation}
    A = \mathbf{M}(\{\mathcal{C}_t\}^T_{t=1})
\end{equation}
The merit of this narrative answer manner is that it explicitly summarizes the reasoning process using text, granting global understanding and harnessing the strong comprehension ability of text-only language models. Moreover, the video language model focuses on the key information and filters out unrelated visual clues across segments.\\
\textbf{Temporal self-refocus for local perception questions.} Unlike global reasoning questions, many long video questions require the model to capture a specific moment. However, due to the stochastic nature of frame sampling, key visual events may be overlooked, leading to suboptimal predictions. To this end, we propose an uncertainty-guided temporal self-refocus method to enhance understanding of the critical moments in long videos. 
The external language model $\mathbf{M}$ first analyzes the question to determine the essential visual cue for the question. For instance, to answer ``What color shoes does the little girl wear?'', $\mathbf{M}$ generates a complementary localization question: ``Is the video showing a little girl with shoes?''. Then we uniformly separate the long video into $T$ segments, $T$ is also 8 here. Then for each segment, the video model $\mathbf{V}$ answers the complementary question based on uniformly sampled 32 frames $\mathcal{F}$. The segment where the video model assigns the highest logit score to the positive token ``Yes'' $l^y$ is identified as the most probable moment containing the target event. 
Once the key moment is localized, the video model produces the complementary answer $A$ based on the frames uniformly sampled from the selected segment, which can be formulated as: 
\begin{equation}
    \quad A = \mathbf{V}(\mathcal{F}_t), \quad t = \mathop{\rm argmax}\limits_{t \in \{1, \dots, T\}} l^y_t.
\end{equation}
This temporal self-refocus mechanism guides the video model to emphasize local comprehension through a self-reflection procedure.
Finally, the complementary answer $A$ is used to compute a binary score $\mathbf{S}^v$ defined as: 
\begin{equation}
    \mathbf{S}^v_p = 
\begin{cases} 
1 & \text{if  } p = A \\
0 & \text{if  } p \neq A ,
\end{cases}
\end{equation}
which is a voting mechanism for the candidates in $\mathcal{P}$.

\textbf{Overall self-reward score.}
We define the final score for prediction selection by a linear combination of the three aforementioned scores $\mathbf{S}^f$, $\mathbf{S}^{mc}$, and $\mathbf{S}^v$:
\begin{equation}
\begin{aligned}
    \mathbf{S}_p &= \mathbf{S}^f_p + \alpha \mathbf{S}^{mc}_p + \beta \mathbf{S}^v_p, \\
    \hat{p}  &=\mathop{\rm argmax}\limits_{p\in \mathbb{\mathcal{P}}} \mathbf{S}_p.
\end{aligned}
\end{equation}
The prediction $\hat{p}$ with the highest overall score $\mathbf{S}$ is selected as the final answer. $\alpha$ and $\beta$ control the weight for $\mathbf{S}^{mc}$ and $\mathbf{S}^f$ respectively, which are manually adjusted. Note that when all predictions in $\mathcal{P}$ reach consensus, the unanimous answer is directly selected. In this case, $\mathbf{S}$ is not computed in our implementation for efficiency.

%% file: sec/5_experiments.tex
\section{Experiments}

\begin{table*}[t]
%\footnotesize
  \centering \scalebox{0.9}{
  \begin{tabular}{cccc|c|c|c|c|c|c} 
  \toprule
  & &   &   \multicolumn{1}{c}{V-MME} & \multicolumn{1}{c}{Long} & \multicolumn{1}{c}{MLVU} & \multicolumn{1}{c}{LVB} & \multicolumn{1}{c}{CG-B$^\blacklozenge$} & \multicolumn{1}{c}{V-MMMU} & \multicolumn{1}{c}{Egoschema} \\
  \cmidrule(lr){4-10}
  \multicolumn{1}{c}{Model} & \multicolumn{1}{c}{F/S} & \multicolumn{1}{c}{Method} & \textit{w/o subs} & \textit{val} & \textit{m-avg} & \textit{test} & \textit{long} & \textit{avg} &  \textit{full}\\  
  \midrule
  \multicolumn{10}{l}{\textit{\small Proprietary models}}\\
  GPT4O  &- &-& 71.9 &66.7 & 64.6 & 34.7 &44.9&61.22 &-\\
  Gemini 1.5 Pro  &- &-& 75.0 &64.0 & - & 33.1 &37.8&53.89 &71.2\\
  Gemini 1.5 Flash  &- &-& 70.3 &61.6 & - & - &33.5&49.78 &66.8\\
  \cmidrule{1-10}
  \multicolumn{10}{l}{\textit{\small Open-source models}}\\
  Qwen2.5-VL-72B  & 768 &-& 73.3 & 60.7 & 74.6 & 47.3 &- & 60.2 & 76.2\\
  NVILA-7B  & 1024 &-& 64.2 & 57.7 & 70.1&-&-&-&-\\
  Oryx-1.5-7B & 128 &-& 58.8 & 56.3 & 67.5&-&-&-&-\\
\cmidrule{1-10}
\multirow{6}{*}{VideoLlama2-7B}  & 16 & \textit{reported} & 47.9 & - & - & - & -& - &  51.7\\
\cmidrule{2-10}
 & \multirow{5}{*}{16} & \textit{baseline}& 47.45& 46.00 & 50.94 &32.73&23.73&25.67 & 51.94 \\
 &  & \textit{Majority}&49.11 &46.37&51.72&32.21&24.33&25.78 & 52.10 \\
 &  &\textbf{Ours}&\textbf{51.04}&\textbf{47.42}&\textbf{55.31}&\textbf{34.22}&\textbf{25.63}&\textbf{27.33} & \textbf{53.85} \\
 &  & $\triangle$ & \textit{\textcolor{red}{\small$+$3.59}} & \textit{\textcolor{red}{\small$+$1.42}} & \textit{\textcolor{red}{\small$+$4.37}} & \textit{\textcolor{red}{\small$+$1.49}} & \textit{\textcolor{red}{\small$+$1.90}}  & \textit{\textcolor{red}{\small$+$1.66}} & \textit{\textcolor{red}{\small$+$1.91}}\\
 &  &\textit{Pass@10}& 74.03 &67.17&74.76&64.36&53.37&45.56 &-\\
\cmidrule{1-10}
\multirow{6}{*}{InternVL-2.5-8B}  & - & \textit{reported}& 64.2 & 60.0 & 68.9 & - & - & - & -\\
\cmidrule{2-10}
 & \multirow{5}{*}{32} & \textit{baseline}& 63.92 &57.51&67.40&42.22&36.63& 46.33 & 65.61\\
 & & \textit{Majority}& 64.19&57.97&68.18 & 43.06 &37.70&46.44 & 66.13\\
 & & \textbf{Ours}&\textbf{65.37}&\textbf{60.66}&\textbf{69.98}& \textbf{46.61} &\textbf{42.17}&\textbf{49.44} & \textbf{67.14}\\
 &  & $\triangle$ & \textit{\textcolor{red}{\small$+$1.45}} & \textit{\textcolor{red}{\small$+$3.15}} & \textit{\textcolor{red}{\small$+$2.58}} & \textit{\textcolor{red}{\small$+$4.39}} & \textit{\textcolor{red}{\small$+$5.54}} & \textit{\textcolor{red}{\small$+$3.11}} & \textit{\textcolor{red}{\small$+$1.53}} \\
 & &\textit{Pass@10}& 76.89&70.68&83.22 &66.50&59.00&54.33 &-\\
\cmidrule{1-10}
\multirow{6}{*}{Llava-Video-7B} & 64 & \textit{reported}& 63.3 & 58.2 & 70.8 & - & - & 36.11 & 57.3 \\
\cmidrule{2-10}
 & \multirow{5}{*}{32} & \textit{baseline}& 62.24 & 57.89& 67.49&41.32&34.70& 37.67 & 62.67 \\
 & & \textit{Majority}&63.59&58.26& 69.37 &42.03 & 35.90& 38.00 & 63.88\\
 & & \textbf{Ours}&\textbf{66.52}&\textbf{61.41}&\textbf{73.38}&\textbf{46.16} &\textbf{39.67}& \textbf{40.44} & \textbf{64.92}\\
 &  & $\triangle$ & \textit{\textcolor{red}{\small $+$4.28}} & \textit{\textcolor{red}{\small$+$3.52}} & \textit{\textcolor{red}{\small$+$5.89}} & \textit{\textcolor{red}{\small$+$4.84}} & \textit{\textcolor{red}{\small$+$4.97}} & \textit{\textcolor{red}{\small$+$2.77}}  & \textit{\textcolor{red}{\small$+$2.25}}\\
 & &\textit{Pass@10}&78.74&71.20&84.28 &69.27 & 56.67 & 46.00 &-\\
\bottomrule
  \end{tabular}}
  \vspace{-2mm}
  \caption{\textbf{Results on long video understanding benchmarks.} All the values in \textit{baseline} lines are the results we reproduced using the public model weights. F/S represents the frame number of each sample. $\blacklozenge$: We do not use the subtitles in CG-Bench. }
  \label{tab:main}
  \vspace{-3mm}
\end{table*}

\begin{figure}[t]
\begin{center}
\vspace{-.04in}
\includegraphics[width=0.95\linewidth]{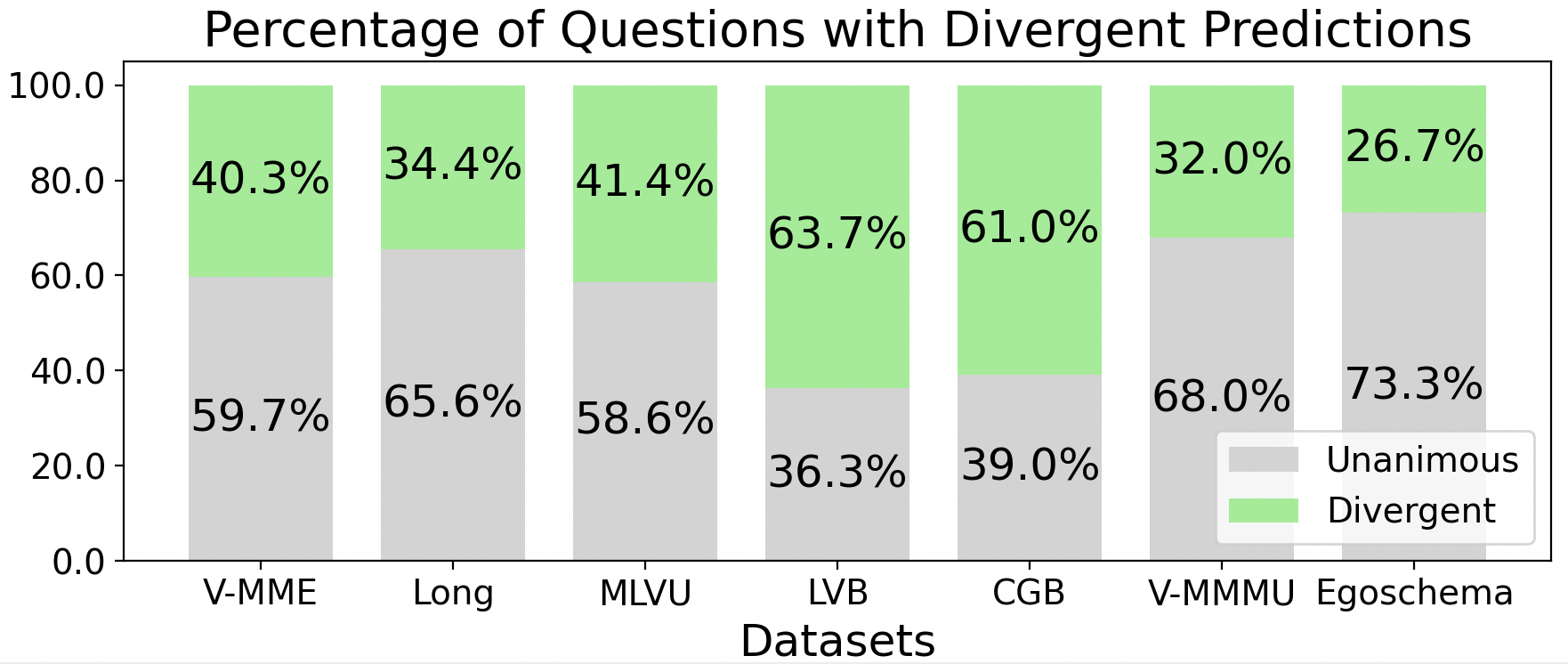}
\end{center}
\vspace{-.25in}
   \caption{Percentage of questions with divergent predictions.}
\label{fig:percentage}
\vspace{-.22in}
\end{figure}

\subsection{Datasets and Setups}
We conduct experiments on the following benchmarks:\\
\textbf{Video-MME} (V-MME) \cite{fu2024video}: a comprehensive benchmark containing various question types. The video duration ranges from a few minutes to over an hour. We report the results under the without subtitle setting.\\
\textbf{LongVideoBench} (Long) \cite{wu2025longvideobench}: manually curated 6,678 referring reasoning questions on videos with an average length of 8 minutes. Validation set results are reported.\\
\textbf{MLVU} \cite{zhou2024mlvu}: a benchmark including multiple tasks, the average video length is about 15 minutes. M-avg is reported.\\
\textbf{LVBench} (LVB) \cite{wang2024lvbench}: an extremely long video benchmark. The average video duration reaches an hour.\\
\textbf{CG-Bench} (CGB) \cite{chen2024cg}: with questions annotated with a temporal clue section. We report the result on the long setting, \ie, not using the temporal annotation. \\
\textbf{Video-MMMU} (V-MMMU) \cite{hu2025video}: evaluates knowledge acquisition ability, we test the average accuracy on all subsets. \\
\textbf{Egoschema} \cite{mangalam2024egoschema}: features 5000 egocentric video question-answer pairs, and the video length is around three minutes. 
\begin{table}[t]
  \centering \scalebox{0.93}{
  \begin{tabular}{c|c|c} 
  \toprule
Dataset &  Random & Ours \\  
\midrule
  V-MME& 35.75\% & \textbf{51.75\%} \\
  Long & 36.96\% & \textbf{50.22\%} \\
  MLVU& 36.89\% & \textbf{62.22\%} \\
  LVB & 30.60\% & \textbf{42.05\%} \\
  CGB& 23.67\% & \textbf{34.23\%} \\
  V-MMMU & 21.88\% & \textbf{31.60\%} \\
\bottomrule
  \end{tabular}}
  \vspace{-1.5mm}
  \caption{Voting accuracy compared with random selection.}
  \label{tab:percentage}
  \vspace{-6mm}
\end{table}
The above datasets cover long videos of different lengths and test the model capability across multiple aspects. As for the video large language models, we use three models:\\
\textbf{VideoLlama2} \cite{cheng2024videollama}: This model uses Clip-large-336 \cite{radford2021learning} as the visual encoder and Mistral-7B-Instruct \cite{jiang2023mistral7b} as the language model, trained on 16 frames.\\
\textbf{InternVL-2.5} \cite{chen2024expanding}: This model adopts InternViT-300M-448px and InternLM2.5-7b-chat as the visual encoder and language model component respectively.\\
\textbf{Llava-Video} \cite{zhang2024video}: Using the Qwen2-7B \cite{wang2024qwen2} language model and the Siglip \cite{zhai2023sigmoid} visual encoder. The model is trained with a frame number of 64.

The above video models are based on various language model architectures and are trained on different numbers of frames, allowing us to evaluate the robustness of the proposed method. 
Regarding the external large language model, we test two open-source state-of-the-art models: Llama-3.3-70B \cite{dubey2024llama} for improved inference efficiency and Deepseek-R1\cite{guo2025deepseek} for superior performance.

\begin{table}[t]
  \centering \scalebox{0.95}{
  \begin{tabular}{l|c|c|c} 
  \toprule
  
 Method & V-MME & Long & MLVU\\  \midrule
 VideoLLama2& 47.45 & 46.00 & 50.94 \\
 $+\mathbf{S}^f$&49.11&46.37& 51.72 \\
 $+\mathbf{S}^f+\mathbf{S}^{mc}$&49.81& 46.89 & 53.98 \\
 $+\mathbf{S}^f+\mathbf{S}^{mc}+\mathbf{S}^v$&51.04 &47.42 & 55.31\\
 \midrule
 Llava-Video& 62.24 & 57.89 & 67.49 \\
 $+\mathbf{S}^f$&63.59&58.26& 69.37 \\
 $+\mathbf{S}^f+\mathbf{S}^{mc}$&64.48& 59.46 & 71.17 \\
 $+\mathbf{S}^f+\mathbf{S}^{mc}+\mathbf{S}^v$& 66.52 & 61.41 & 73.38\\
\bottomrule
  \end{tabular}}
  \vspace{-2mm}
  \caption{Ablation on different scores.}
  \label{tab:components}
  \vspace{-3.5mm}
\end{table}

\subsection{Implementation Details}
The specific versions of video models used in the experiments are VideoLLaMA2-7B-16F, InternVL2-5-8B, and LLaVA-Video-7B-Qwen2. As for the Llama-3.3-70B language model, we adopt the 8-bit quantized version. For efficient attention computation, we adopt Scaled Dot Product Attention (SDPA) in Pytorch. In CG-Bench, subtitles are not included to reflect the influence of the visual context. Prompts for the external language model are provided in the supplementary material. All evaluations are based on lmms-eval \cite{zhang2024lmmsevalrealitycheckevaluation} repository. $\alpha$ and $\beta$ are set to 1 and 3 by default and are slightly adjusted according to the dataset.
\subsection{Quantitative Results}
\textbf{Main results.} \cref{tab:main} reports the results on three models and seven long video benchmarks. The models have distinct language model architectures and are trained on different frame numbers, evaluating the robustness of our method. 
The \textit{reported} row copies the values reported in the original paper. However, since some inference settings and implementation details are unavailable, we reproduce the results of each model as our baseline using the official model checkpoints with uniformly sampled frames. 
For VideoLlama2, each sample contains 16 frames following the original setting, while InternVL-2.5 and Llava-Video use 32 frames to reduce memory consumption and improve inference speed. A simple majority answer selection strategy outperforms the baseline method in most cases. Our method further surpasses both the baseline and majority voting across three models. Compared to vanilla Llava-video, our method achieves a $+4.28\%$ improvement on Video-MME, $+5.89\%$ on MLVU, and $+4.97\%$ on CG-Bench. When combined with InternVL-2.5, the accuracy boosts to 42.17\% on CG-Bench and 46.61\% on LVBench, comparable with SOTA proprietary models. In terms of VideoLlama2, our method also improves performance across all benchmarks, though the gains are smaller compared to Llava-Video and InternVL-2.5. One reason is that each sample in VideoLLaMA2 only contains 16 frames, limiting the ability to perceive the full video content. Moreover, we observe that the narration and instruction-following capabilities of VideoLlama2 are slightly inferior to the other two models, increasing difficulty in obtaining video information and implicit key moment locating. 
In essence, the proposed method effectively narrows the gap between accuracy and Pass@10. However, there is still room for improvement, indicating the potential of scaling visual context sampling for long video understanding.\\
\begin{table}[t]
  \centering \scalebox{0.95}{
  \begin{tabular}{cc|c|c|c} 
  \toprule
 Model &Method & V-MME & Long & MLVU\\  \midrule
 \multirow{4}{*}{VideoLLama2} &\textit{baseline}& 47.45 & 46.00 & 50.94 \\
 &\textit{Majority}&49.11&46.37& 51.72 \\
 &\textbf{Ours-L}& 50.11 & 47.34 & 55.13 \\
 & \textbf{Ours-D}& 51.04 & 47.42 & 55.31\\
 \midrule
 \multirow{4}{*}{Llava-Video} &\textit{baseline}& 62.24 & 57.89 & 67.49 \\
 &\textit{Majority}&63.59&58.26& 69.37 \\
 &\textbf{Ours-L}&66.30&60.73& 72.37 \\
 & \textbf{Ours-D}&66.52 & 61.41 & 73.38\\
\bottomrule
  \end{tabular}}
  \vspace{-2mm}
  \caption{Ablation on large language models. Ours-L uses Llama3.3-70B model while Ours-D uses Deepseek-R1.}
  \label{tab:external}
  \vspace{-5mm}
\end{table}
\textbf{Dataset statistics.} As mentioned in \cref{sec:type}, the score for prediction selection is only calculated when there are disagreements among predictions. We analyze how many questions in each dataset require calculate the score and report the percentage of such questions relative to the entire dataset in \cref{fig:percentage}. Across seven datasets, the average percentage is about $40\%$, though with notable variance. Only $26.7\%$ questions in Egoschema have divergent answers, likely because videos in Egoschema are short, thereby the divination in visual contexts is not drastic. In contrast, for longer benchmarks like LVBench and CG-Bench, the percentage increases to around $60\%$. This is because each sample only views 32 frames from hour-long videos, leading to greater uncertainty as video length increases. 
We also report the percentage of correct prediction selection on each benchmark in \cref{tab:percentage}. To demonstrate the effectiveness of our method, we compare the accuracy with a baseline that randomly picks a prediction. Our method achieves a significant improvement of $+12.3\%$ on average. However, there is still room for improvement, highlighting the potential for further work on enhancing the selection of the correct answer among predictions. 
\subsection{Ablation Studies}
\textbf{Effectiveness of different components.} \cref{tab:components} presents an ablation study on the effectiveness of each proposed score on three datasets (Video-MME, Longvideobench, and MLVU) using two models (Llava-Video and VideoLlama2). Applying only $S^f$ to the baseline method is equivalent to selecting the majority answer, improving the accuracy. Incorporating $\mathbf{S}^{mc}$ further enhances performance by approximately $1\%$, while adding $\mathbf{S}^v$ achieves the best results, yielding an average increase of $2\%$. Each component consistently contributes to performance gains across all benchmarks and models, demonstrating its effectiveness. \\
\begin{figure}[t]
\begin{center}
\includegraphics[width=0.93\linewidth]{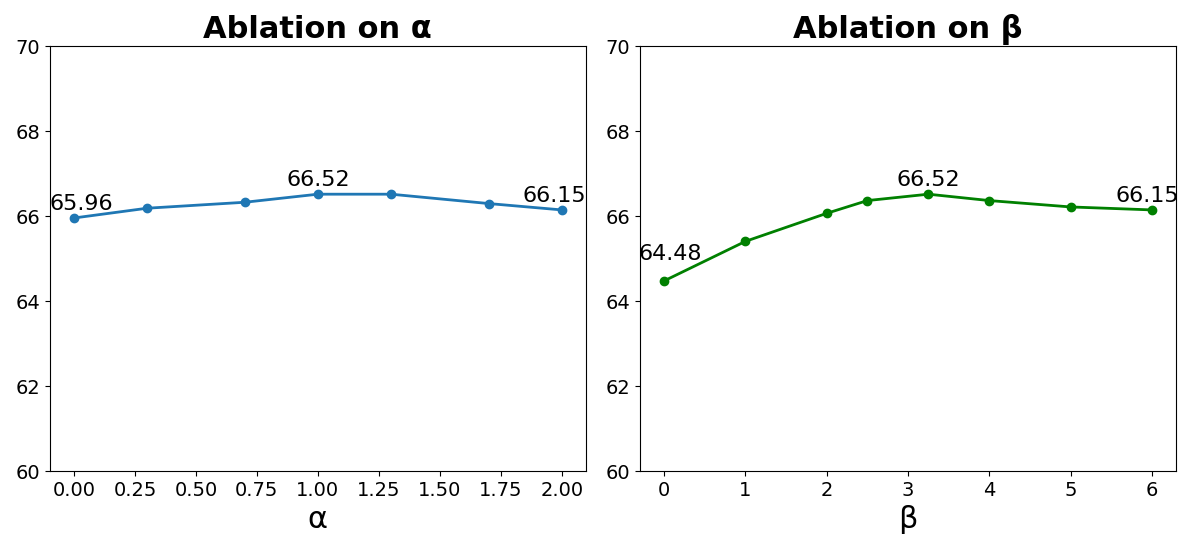}
\end{center}
\vspace{-.3in}
   \caption{Influence of $\alpha$ and $\beta$ using Llava-video on Video-MME.}
\label{fig:alphabeta}
\vspace{-.1in}
\end{figure}
\begin{table}[t]
  \centering \scalebox{0.87}{
  \begin{tabular}{cc|c|c|c} 
  \toprule
  
Relation & Method & V-MME & Long & MLVU\\  
\midrule
  \multirow{2}{*}{Intra}&Max & 66.26 & 61.25 & 73.01 \\
  &\textbf{Marginal} &\textbf{66.52} & \textbf{61.41} & \textbf{73.38} \\
\midrule
  \multirow{2}{*}{Inter}&Mean & 66.22 & 61.26 & 73.28 \\
 &\textbf{Max} & \textbf{66.52}& \textbf{61.41} & \textbf{73.38} \\
\bottomrule
  \end{tabular}}
  \vspace{-2mm}
  \caption{Comparison of confidence types using Llava-Video.}
  \label{tab:confidence}
  \vspace{-4mm}
\end{table}
\textbf{Ablation on an external language model.} In \cref{tab:external}, we replace the external language model used for question categorization, substituting Deepseek-R1 \cite{guo2025deepseek} with Llama-3.3-70B \cite{dubey2024llama} 8-bit quantized version. This ablation evaluates the impact of the language model on performance. We report the accuracy using Llama3.3 in the Ours-L rows and the accuracy using Deepseek-R1 in the Ours-D rows. 
We test two base models, \ie, VideoLlama2 and Llava-Video, on three benchmarks including Video-MME, Longvideobench, and MLVU. The results indicate a slight performance degradation when using Llama3.3-70B. 
For instance, accuracy decreases by $0.2\%$ on Video-MME with Llava-Video, and by  $0.08\%$ and $0.19\%$ on LongVideoBench and MLVU respectively with VideoLlama2. These findings suggest that a stronger language model contributes to higher accuracy. Moreover, despite the slight drop in performance, using Llama3.3-70B still outperforms the majority answer, showing the robustness toward different language models.\\
\textbf{Influence of score weights $\alpha$ and $\beta$.}\cref{fig:alphabeta} illustrates the relationship between accuracy and the weight values $\alpha$ and $\beta$ on the Video-MME using Llava-Video. Performance drops when setting $\alpha$ or $\beta$ to 0, highlighting the importance of both scores. However, over-high values of $\alpha$ and $\beta$ also lead to suboptimal performance. Overall, accuracy remains stable across various weights, demonstrating the robustness the proposed method to $\alpha$ and $\beta$.\\
\textbf{Comparing different confidence designs.} In \cref{tab:confidence}, we explore different strategies for aggregating inter-intra sample confidence on the Video-MME using Llava-Video. 
For intra-sample confidence, we replace the marginal confidence score with the logit value of the most probable token, resulting in a slight performance degradation. This demonstrates the effectiveness of estimating logit differences to compare confidence across tokens. 
As for inter-sample confidence, we replace the max value with the mean value of each type of prediction. Results show that using the max value achieves higher accuracy, as it effectively filters out predictions with high uncertainty.\\
\textbf{Influence of sample frames.} We investigate how the number of frames per sample affects performance, as a smaller frame number leads to faster inference.  \cref{tab:framenum} shows the accuracy of various frame numbers on three benchmarks using Llava-Video. When reducing the frame number to 8, the baseline accuracy on Video-MME drops significantly to $56.26\%$, whereas our method achieves $62.48\%$, maintaining a $+6.22\%$ improvement. However, with 16 frames, the improvement over the baseline decreases to $+3.27\%$ on Video-MME. 
This trend indicates that our method consistently improves the performance and the improvement becomes more pronounced as the frame number decreases. A similar pattern also appeared in the other two benchmarks, suggesting a potential approach to improve inference speed by reducing the number of frames per sample.

%% file: sec/6_conclusion.tex
\section{Conclusion}
\begin{table}[t]
  \centering \scalebox{0.87}{
  \begin{tabular}{cc|c|c|c} 
  \toprule
 Frames &Method & V-MME & Long & MLVU \\  \midrule
 \multirow{4}{*}{8} &\textit{baseline}& 56.26 & 55.65 & 59.08 \\
 &\textit{Majority}& 58.44& 55.95 & 60.64 \\
 & \textbf{Ours}& \textbf{62.48} &\textbf{59.46}& \textbf{66.94} \\
 & \textit{Pass@10}&78.48&73.59& 81.93\\
 \midrule
 \multirow{4}{*}{16} &\textit{baseline}& 60.18 & 56.69 & 62.94 \\
 &\textit{Majority}&60.74 & 57.44 & 65.05 \\
 & \textbf{Ours}& \textbf{63.45} & \textbf{60.73}& \textbf{70.48}\\
 & \textit{Pass@10}& 78.18 & 72.02& 82.98\\
 \midrule
 32 &\textbf{Ours}& \textbf{66.52} & \textbf{61.41} & \textbf{73.38} \\
\bottomrule
  \end{tabular}}
  \vspace{-2mm}
  \caption{Ablation on frame numbers using Llava-Video.}
  \label{tab:framenum}
  \vspace{-5mm}
\end{table}
In this paper, we propose to improve long video understanding by sampling multiple predictions and selecting the optimal one. Our approach is inspired by the finding that video models exhibit high coverage when provided with diverse visual contexts as input. To identify the best prediction, we devise a score combining frequency score, marginal confidence score, and adaptive voting score. Experiments on seven benchmarks and three models demonstrate the effectiveness of our method. We hope this work inspires future research on visual sampling in video tasks.